%% file: main.tex
\documentclass[dvipsnames]{article}
\input{preamble}

\begin{document}
\input{paper_body}
\end{document}

%% file: preamble.tex
\usepackage{frontierchallenge_conference}

\usepackage{iftex}
\ifPDFTeX
  \usepackage[utf8]{inputenc}
  \usepackage[T1]{fontenc}
\else
  \usepackage{fontspec}
\fi
\usepackage{amsmath,amssymb,amsfonts}
\usepackage{booktabs}
\usepackage{array}
\usepackage{multirow}
\usepackage{makecell}
\usepackage{tabularx}
\usepackage{graphicx}
\usepackage{subcaption}
\usepackage{placeins}
\usepackage{float}
\usepackage[all]{hypcap}
\usepackage{setspace}
\usepackage{ragged2e}
\usepackage{enumitem}
\usepackage{tcolorbox}

\input{macro.tex}

\title{FrontierChallenge: Evaluating Scientific Workflow Completion}
\author{\normalsize Apodex Team\thanks{The full contributor list is provided
in Appendix~\ref{app:contributors}.}}

%% file: macro.tex
\newcommand{\benchmark}{\textsc{FrontierChallenge}}

\newcommand{\cmark}{\tikz[baseline=-0.55ex]{\fill[black!78] (0,0) circle (1.7pt);}}
\newcommand{\xmark}{\textcolor{black!30}{\rule[0.55ex]{0.62em}{0.45pt}}}

\newif\iffrontierempiricalresults
\newif\iffrontiermockresults
\newif\iffrontierjudgeresults
\newif\iffrontierevaluatorresults
\newif\iffrontierevaluatorfigureresults
\newif\iffrontierrobustnessresults
\newif\iffrontierharnessresults
\newif\iffrontierfailureresults
\newif\iffrontierartifactresults
\frontierempiricalresultsfalse
\frontiermockresultsfalse
\frontierjudgeresultsfalse
\frontierevaluatorresultsfalse
\frontierevaluatorfigureresultsfalse
\frontierrobustnessresultsfalse
\frontierharnessresultsfalse
\frontierfailureresultsfalse
\frontierartifactresultsfalse
\InputIfFileExists{tables/results_mode.tex}{}{\frontierempiricalresultsfalse}
\InputIfFileExists{tables/judge_results_mode.tex}{}{\frontierjudgeresultsfalse}
\InputIfFileExists{tables/evaluator_results_mode.tex}{}{\frontierevaluatorresultsfalse}
\InputIfFileExists{tables/evaluator_figure_results_mode.tex}{}{\frontierevaluatorfigureresultsfalse}
\InputIfFileExists{tables/robustness_results_mode.tex}{}{\frontierrobustnessresultsfalse}
\InputIfFileExists{tables/harness_results_mode.tex}{}{\frontierharnessresultsfalse}
\InputIfFileExists{tables/failure_results_mode.tex}{}{\frontierfailureresultsfalse}
\InputIfFileExists{tables/artifact_results_mode.tex}{}{\frontierartifactresultsfalse}

\newcommand{\modelIcon}[1]{\raisebox{-0.18em}{\csname frontiermodelicon#1\endcsname}\hspace{0.35em}}

\newcommand{\frontierwebicon}{\raisebox{-1.5pt}{\includegraphics[height=1.15em]{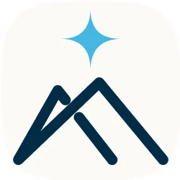}}}
\newcommand{\frontiergithubicon}{\raisebox{-1.5pt}{\includegraphics[height=1.05em]{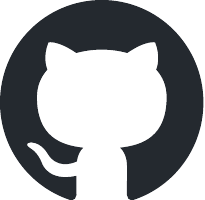}}}
\newcommand{\frontierhuggingfaceicon}{\raisebox{-1.5pt}{\includegraphics[height=1.15em]{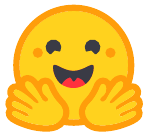}}}
\newcommand{\frontierprojectresource}[3]{\href{#1}{#2\enspace\textbf{#3}}}
\newcommand{\frontierprojectlinks}{%
  \begin{center}
    \vspace{-0.6em}
    \small\rmfamily
    \begin{tabular}{@{}c@{\hspace{1.8em}}c@{\hspace{1.8em}}c@{}}
      \frontierprojectresource{https://apodexai.github.io/FrontierAgent/benchmarks/FrontierChallenge/}{\frontierwebicon}{Website (Leaderboard)} &
      \frontierprojectresource{https://github.com/ApodexAI/FrontierAgent/tree/main/benchmarks/frontierchallenge}{\frontiergithubicon}{GitHub (Evaluation)} &
      \frontierprojectresource{https://huggingface.co/datasets/apodex/FrontierChallenge}{\frontierhuggingfaceicon}{Hugging Face (Dataset)}
    \end{tabular}
    \vspace{0.2em}
  \end{center}%
}

\newtcolorbox{writingguide}[1][]{
  colback=gray!6,
  colframe=gray!35,
  boxrule=0.4pt,
  arc=1mm,
  left=3mm,
  right=3mm,
  top=2mm,
  bottom=2mm,
  fonttitle=\bfseries,
  title={Writing guide},
  #1
}

%% file: tables/results_mode.tex
\frontierempiricalresultsfalse

%% file: tables/judge_results_mode.tex
\frontierjudgeresultsfalse

%% file: tables/evaluator_results_mode.tex
\frontierevaluatorresultsfalse

%% file: tables/evaluator_figure_results_mode.tex
\frontierevaluatorfigureresultsfalse

%% file: tables/robustness_results_mode.tex
\frontierrobustnessresultsfalse

%% file: tables/harness_results_mode.tex
\frontierharnessresultsfalse

%% file: tables/failure_results_mode.tex
\frontierfailureresultsfalse

%% file: tables/artifact_results_mode.tex
\frontierartifactresultsfalse

%% file: paper_body.tex
\maketitle
\frontierprojectlinks
\iffrontiermockresults
\begin{center}
  \begin{tcolorbox}[
    colback=BrickRed!5,
    colframe=BrickRed!75,
    boxrule=0.8pt,
    arc=0.8mm,
    left=2mm,right=2mm,top=1.2mm,bottom=1.2mm,
    width=0.94\textwidth]
    \centering\bfseries\color{BrickRed}
    MOCK RESULTS PREVIEW. EVERY NUMERICAL RESULT IS SYNTHETIC AND MUST NOT BE
    CITED, RELEASED AS A LEADERBOARD, OR USED FOR SUBMISSION.
  \end{tcolorbox}
\end{center}
\fi
\input{sections/abstract}

\input{sections/introduction}
\input{sections/related_work}
\input{sections/benchmark_design}
\input{sections/evaluation_framework}
\input{sections/experiments}
\input{sections/conclusion}

\begingroup
\small
\bibliographystyle{frontierchallenge_template}
\phantomsection
\bibliography{main}
\endgroup

\clearpage
\appendix
\section*{Appendix}
\input{sections/appendix}

%% file: sections/abstract.tex
\begin{abstract}
Scientific agents increasingly analyze data, execute code, and produce research
artifacts, yet most benchmarks emphasize final answers, isolated programs, or a
single domain. We introduce \benchmark{}, a cross-domain benchmark comprising
300 end-to-end scientific workflows. In this paper, we release and evaluate 97
of these tasks, spanning quantum chemistry, molecular dynamics, materials
characterization, analytical chemistry, life science, and
electrochemistry/environment.
Each task provides fixed inputs and specifies a bundle of required scientific
deliverables. We evaluate twelve frontier models with three agent scaffolds.
Pass Rate measures the fraction of tasks satisfying the
full-completion criterion, while Avg.\ Score captures partial progress. 
Each of the best-performing configurations completed only 20 of the 97
released tasks, yielding a
Pass Rate of 20.6\%. Partial progress translated especially poorly into complete
delivery in analytical chemistry and electrochemistry/environment: Avg.\ Scores
reached
87.6 and 94.9, but the highest Pass Rates were only 4\% and 0\%. Among
non-passing Claude Code trajectories, 75.5\% still ended with language claiming
completion. Complementary HDS6 process scores correlate strongly with task
outcomes, supporting \benchmark{} as a benchmark of Heavy Duty Solver
capabilities. These findings show that neither high partial scores nor confident
claims of completion reliably indicate that a scientific task has been fully
delivered, highlighting the need to evaluate end-to-end workflow execution and
the completeness of scientific deliverables together.
\end{abstract}

%% file: sections/introduction.tex
\section{Introduction}\label{sec:introduction}

Language models are evolving from text generators into agents that can plan,
call tools, execute code, and modify persistent files
\citep{liu2024agentbench,mialon2024gaia,xie2024osworld}. Alongside advances in
agent scaffolding, continual pre-training has been explored as a way to scale
general agent capabilities \citep{su2025scaling}. Recent systems suggest
that agentic support can extend beyond isolated tasks such as literature
retrieval or text translation to coordinating multi-stage research workflows
with inspectable outputs \citep{lu2024ai}. This shift changes
what constitutes success on a scientific task. Producing a plausible conclusion
is not enough: an agent may need to inspect heterogeneous inputs, select and run
an analysis, validate intermediate results, and deliver mutually consistent
code, tables, figures, and prose.

Existing benchmarks cover expert knowledge, general tool use, software
interaction, code repair, paper replication, and scientific data analysis
\citep{phan2025humanity,jimenez2024swebench,chen2025scienceagentbench,siegel2024core,starace2025paperbench}.
Their evaluation units, however, are often a final answer, an interaction
trace, a single program, or a workflow from one discipline. These settings do
not fully characterize whether an agent can complete heterogeneous scientific
work whose success depends on several analytical stages and several required
deliverables.

\benchmark{} therefore asks a focused question:
\begin{quote}
\itshape Given a specified scientific task and fixed data, can an agent
independently complete the workflow from input processing to final deliverables
and satisfy the complete task contract?
\end{quote}
Reliable execution underpins a trustworthy scientific handoff: analyses and
outputs must be inspectable, reproducible, and mutually consistent.
\benchmark{} therefore targets a narrower capability than autonomous science:
executing a specified scientific workflow after its objective, inputs, and
required outputs are fixed. The agent is not asked to set the research agenda
or formulate the problem.

We constructed a pool of 300 end-to-end scientific workflows grouped into six
reporting domains: quantum chemistry, molecular dynamics, materials
characterization, analytical chemistry, life science, and
electrochemistry/environment. In
this study, we evaluate and publicly release 97 tasks; the remaining 203 are
retained as an internal held-out set. Representative tasks require agents to
use domain software, including ORCA, CP2K, LAMMPS, AmberTools, and PLUMED;
perform quantitative analysis, quality control, and visualization; and produce
executable code and evidence-grounded reports. The unit of evaluation is the
complete submitted artifact bundle rather than a single answer.

We evaluated twelve frontier models with three agent scaffolds on the 97 tasks.
Our primary metric
is Pass Rate, defined as the fraction of
tasks for which the complete task contract is satisfied. Avg.\ Score is
reported only as a complementary measure of partial progress; a high-scoring
but incomplete submission is not counted as a pass.

To complement outcome-based evaluation, we examine how agents carry out these
workflows. Following the HDS6 framework introduced in \emph{Apodex
Discovery}~\citep{wang2026apodexdiscovery}, we assess Coherence, Evidence,
Alternatives, Scope, Tools, and Repair in 1,164 frozen trajectories from
12 model--scaffold configurations. HDS6 process scores correlate strongly
with both Avg.\ Score and Pass Rate (Section~\ref{sec:hds6}), providing
convergent evidence that \benchmark{} tests Heavy Duty Solver capabilities
through the execution and delivery of scientific workflows.

The central finding is a persistent gap between partial progress and complete
scientific delivery. The best-performing configurations (i.e., GPT-5.6
Sol with Codex and Grok 4.6 with Claude Code) completed only 20 of 97 tasks,
corresponding to a Pass Rate of 20.6\%, despite the highest Avg.\ Score
reaching 87.9. This gap was especially pronounced in analytical chemistry and
electrochemistry/environment: Avg.\ Scores reached 87.6 and
94.9, respectively, while the
highest Pass Rates were only 4\% and 0\%. Thus, although frontier
systems can make substantial progress on complex scientific tasks, they remain
far from reliable end-to-end execution.

This work makes three contributions:
\begin{enumerate}[leftmargin=*,itemsep=2pt,topsep=3pt]
  \item \textbf{Cross-domain scientific workflow benchmark.} We construct a
  pool of 300 end-to-end workflows spanning six reporting domains, evaluate and
  publicly release 97 tasks, and retain 203 as an internal held-out set. Each
  task is defined by fixed inputs, a heterogeneous deliverable contract, and a
  task-specific Grader.
  \item \textbf{Contract-level evaluation.} We use Pass Rate as the
  primary metric of complete workflow execution and Avg.\ Score to
  characterize partial progress across task-specific scientific rubrics.
  \item \textbf{Frontier model study and failure analysis.} We evaluate twelve
  frontier models with multiple agent scaffolds, quantify domain-level
  completion gaps, compare machine-observable contract-breach signatures across
  domains, and analyze final-status and error behavior in 970 Claude Code
  trajectories.
\end{enumerate}

%% file: sections/related_work.tex
\section{Related Work}\label{sec:related-work}

\paragraph{General capability and long-horizon agent benchmarks.}
Humanity's Last Exam (HLE) probes the limits of expert knowledge with difficult,
checkable questions \citep{phan2025humanity}. AgentBench, GAIA, OSWorld, and
SWE-bench extend evaluation to tool use, computer interaction, and software
engineering \citep{liu2024agentbench,mialon2024gaia,xie2024osworld,jimenez2024swebench}.
Agents' Last Exam (ALE) evaluates long-horizon professional work, while
Frontier-Bench collects difficult, verifiable tasks near the frontier of agent
capability \citep{sun2026agentslastexam,harbor2026frontierbench}. These
benchmarks establish the importance of evaluating completed work, but they are
not centered on cross-domain scientific workflows.

\paragraph{Scientific knowledge and data analysis.}
LAB-Bench evaluates knowledge and reasoning required for biological research
\citep{laurent2024labbench}. BixBench, BioMysteryBench, and CompBioBench use
biological data to construct open-ended or objectively verifiable problems
\citep{mitchener2025bixbench,anthropic2026biomysterybench,nair2026agentic}.
BLADE represents open-ended data analysis through expert-recognized analytical
decisions, while ScienceAgentBench asks agents to produce executable analyses
derived from the scientific literature \citep{gu2024blade,chen2025scienceagentbench}.
These benchmarks substantially increase scientific realism, but many evaluate
an answer, a decision set, or a single self-contained program rather than a
heterogeneous deliverable set.

\paragraph{End-to-end scientific workflows.}
CORE-Bench and PaperBench evaluate computational reproduction and full research
replication \citep{siegel2024core,starace2025paperbench}. ScienceBoard,
SciAgentArena, and SciAgentGym assess scientific software use, interactive
research environments, and multi-step tool use
\citep{sun2026scienceboard,liu2026benchmarking,shen2026sciagentgym}.
BioAgent Bench and BiomniBench evaluate end-to-end artifacts or processes in
bioinformatics and biomedicine \citep{fa2026bioagent,qu2026biomnibench}.
NatureBench and AstaBench broaden the scope to paper-level methods and
cross-domain research tasks \citep{wang2026naturebench,bragg2025astabench}.
\benchmark{} is complementary: it focuses on reliable completion after the
scientific objective and inputs are fixed, while combining cross-domain
coverage, multi-artifact outputs, and task-specific executable evaluation.

\begin{table}[t!]
  \centering
  \scriptsize
  \caption{Core design characteristics of related benchmarks. A solid
  dot indicates that the characteristic is part of the benchmark's primary
  design; a dash indicates otherwise.}
  \label{tab:related-comparison}
  \setlength{\tabcolsep}{2.4pt}
  \begin{tabularx}{\textwidth}{@{}>{\RaggedRight\arraybackslash}p{0.19\textwidth}*{6}{>{\centering\arraybackslash}X}@{}}
    \toprule
    \textbf{Benchmark} & \makecell{Scientific\\workflow core} & \makecell{Fixed scientific\\inputs} & \makecell{End-to-end\\execution} & \makecell{Multiple\\artifacts} & \makecell{Executable\\evaluation} & \makecell{Cross-domain\\science} \\
    \midrule
    HLE \citep{phan2025humanity} & \xmark & \xmark & \xmark & \xmark & \xmark & \cmark \\
    ALE \citep{sun2026agentslastexam} & \xmark & \xmark & \cmark & \cmark & \cmark & \xmark \\
    Frontier-Bench \citep{harbor2026frontierbench} & \xmark & \xmark & \cmark & \cmark & \cmark & \xmark \\
    BLADE \citep{gu2024blade} & \cmark & \cmark & \xmark & \xmark & \cmark & \cmark \\
    ScienceAgentBench \citep{chen2025scienceagentbench} & \cmark & \cmark & \cmark & \xmark & \cmark & \cmark \\
    CORE-Bench \citep{siegel2024core} & \cmark & \cmark & \cmark & \cmark & \cmark & \cmark \\
    PaperBench \citep{starace2025paperbench} & \cmark & \cmark & \cmark & \cmark & \cmark & \xmark \\
    BioAgent Bench \citep{fa2026bioagent} & \cmark & \cmark & \cmark & \cmark & \cmark & \xmark \\
    BiomniBench \citep{qu2026biomnibench} & \cmark & \cmark & \cmark & \cmark & \cmark & \xmark \\
    NatureBench \citep{wang2026naturebench} & \cmark & \cmark & \cmark & \cmark & \cmark & \cmark \\
    AstaBench \citep{bragg2025astabench} & \cmark & \cmark & \cmark & \xmark & \cmark & \cmark \\
    \textbf{\benchmark{}} & \cmark & \cmark & \cmark & \cmark & \cmark & \cmark \\
    \bottomrule
  \end{tabularx}
\end{table}

%% file: sections/benchmark_design.tex
\section{FrontierChallenge}\label{sec:design}

\subsection{Task Collection, Curation, and Packaging}\label{sec:tasks}

\paragraph{Task collection.}
We collected tasks from scientific and engineering settings, focusing on
realistic workflows that require domain knowledge, specialized software,
experimental data, or engineering environments. The tasks originate from
analysis, computation, simulation, and research-delivery processes performed in
domain practice, rather than from expanded question answering or isolated
coding exercises. We prioritized workflows in which an agent must understand a
professional objective, use domain tools correctly, execute interdependent
steps, and produce verifiable scientific artifacts. The complete collection
contains 300 scientific workflows; in this paper, we release and evaluate 97.

\paragraph{Screening and quality control.}
We curated tasks according to four design principles:

\begin{itemize}[leftmargin=*,itemsep=2pt,topsep=3pt]
  \item \textbf{Representativeness.} The workflow, software, and methods must
  reflect plausible professional practice.
  \item \textbf{Complexity.} The task must require an end-to-end,
  dependency-aware process culminating in a substantive deliverable, rather
  than a single command, tool call, or local edit.
  \item \textbf{Diversity.} The collection must vary in scientific knowledge,
  workflow type, and difficulty, rather than repeat templates with only inputs
  or parameters changed.
  \item \textbf{Verifiability.} Outputs must be assessable through files,
  numerical values, quantitative measures, or explicit acceptance criteria,
  thereby supporting repeatable automated evaluation.
\end{itemize}

\begin{figure}[t]
  \centering
  \includegraphics[width=\textwidth]{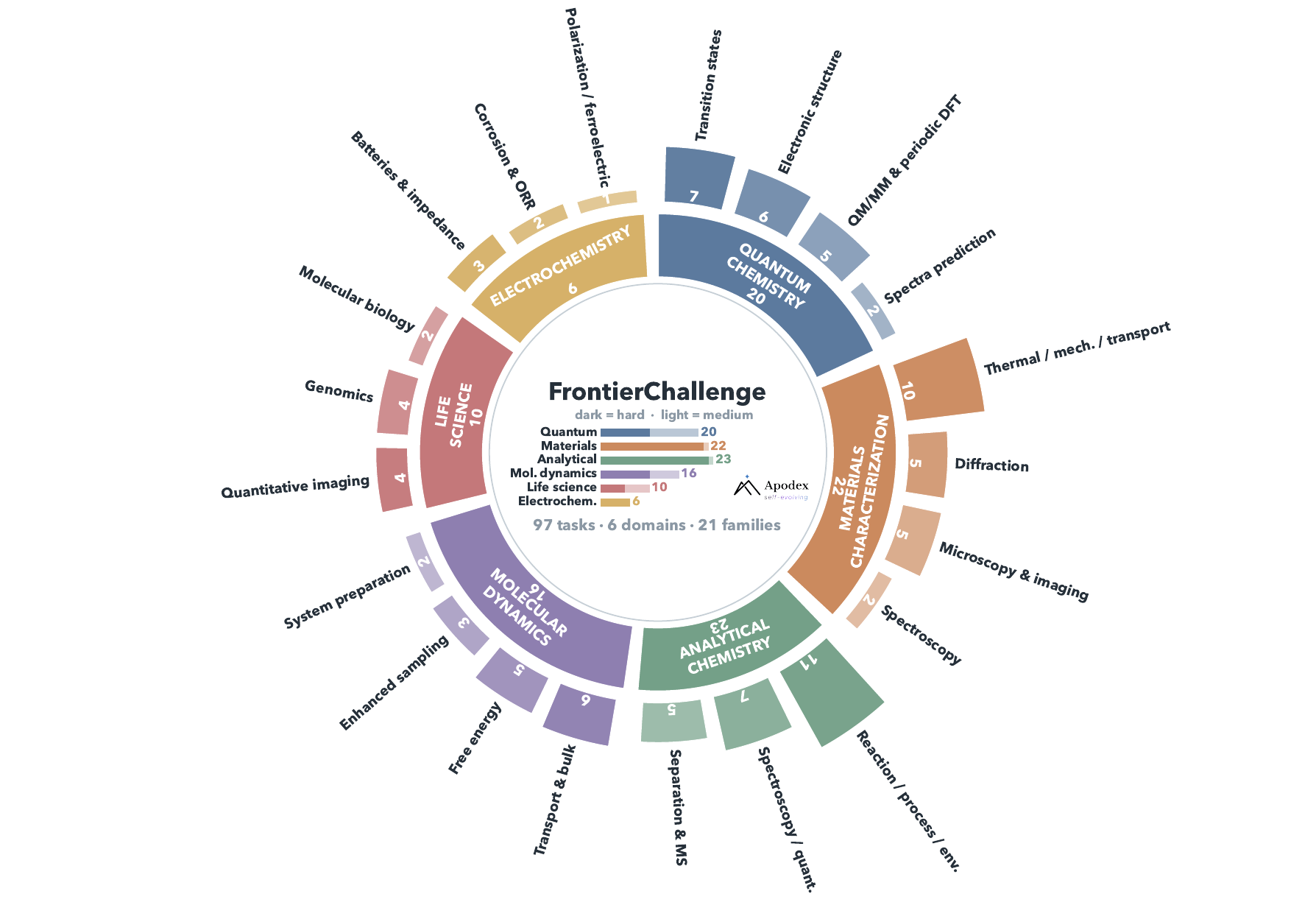}
  \caption{Domain and workflow-family composition of the 97 released tasks;
  center bars show the retained Hard/Medium split.}
  \label{fig:domain-coverage}
\end{figure}
\FloatBarrier

We also checked that every included task had fixed inputs, a defined execution
environment, a complete deliverable contract, and an executable evaluation
procedure. Tasks centered on isolated facts or single-step operations, tasks
with purely subjective outputs, and tasks lacking the materials needed for
reproducible scoring were excluded before the final 300-workflow collection
was formed.

\paragraph{Standardization and packaging.}
After curation, we organized each task as a self-contained package with five
aligned elements: a \emph{task description} defining the scientific objective;
fixed \emph{inputs}, including data and necessary context; the \emph{software
and tools} available in the execution environment; an \emph{output contract}
listing the required deliverables; and an \emph{evaluation procedure} defining
successful completion. Deliverables may include scientific reports, structured
tables, diagnostic figures, executable analysis code, simulation products, or
multiple artifacts that must remain mutually consistent. Completion therefore
depends on the entire artifact bundle, not on whether the agent returns a
plausible final answer.

Each package contains task metadata, agent-facing instructions, input data,
expected-output or reference material, a stepwise scoring rubric, an executable
Grader, and documentation for reproduction and scoring. Environment
specifications, domain tools, auxiliary references, and execution traces are
included when required by the workflow. Agent-visible instructions and inputs
are separated from evaluator-side references and scoring components. This
standardized organization allows each task to be rerun under fixed conditions
and scored consistently across evaluated systems.

\subsection{Dataset Statistics}\label{sec:statistics}

All 300 workflows in the curated collection passed the quality-control checks
above. For this study, we randomly selected 97 tasks from the subset whose
official evaluation does not require GPU resources, publicly released them,
and used them for the reported experiments. The remaining 203 tasks form an
internal held-out set, which also contains workflows whose official evaluation
requires GPUs; none of these held-out tasks is included in the results reported
in this paper. The released tasks comprise 74 Hard tasks and 23 Medium tasks.
They begin from
fixed public empirical data, public sequence or structural resources, or
scientifically constrained synthetic data. These source types describe the
inputs used to instantiate the workflows; they do not imply that every task
begins from a direct laboratory measurement.

For analysis, we organize the 97 released tasks into six reporting domains:
quantum chemistry (20 tasks), molecular dynamics (16), materials
characterization (22), analytical chemistry (23), life science (10), and
electrochemistry (6). Collectively, these tasks cover 21 workflow families and
require heterogeneous deliverables, including scientific reports, structured
data, figures, executable code, and simulation products. Figure~\ref{fig:domain-coverage}
summarizes the domain counts, all 21 workflow families, and the within-domain
Hard/Medium composition.

The six domains are descriptive slices of the released evaluation set, not
probability samples of their corresponding scientific fields. Their sizes
reflect the composition of the current release rather than a claim of balanced
coverage. Accordingly, differences in model performance across domains should
not be interpreted as intrinsic rankings of disciplinary difficulty.

%% file: sections/evaluation_framework.tex
\section{Experiments}\label{sec:experiments}

We conducted experiments to address four research questions:
\begin{enumerate}[leftmargin=*,itemsep=2pt,topsep=3pt]
  \item \textbf{RQ1:} How reliably can current frontier models complete
  specified scientific workflows?
  \item \textbf{RQ2:} How does scientific workflow performance vary across
  domains?
  \item \textbf{RQ3:} Which observable contract breaches and trajectory
  behaviors most often accompany incomplete scientific handoffs?
  \item \textbf{RQ4:} What process capabilities do frozen trajectories reveal
  under a Heavy Duty Solver assessment?
\end{enumerate}

\subsection{Agent Scaffolds and Models}\label{sec:systems}

\paragraph{Agent scaffolds.}
We used three advanced agent scaffolds. These were
Codex~\citep{openai2026codexcli}, Claude
Code~\citep{anthropic2026claudecode}, and Frontier
Agent~\citep{apodexai2026frontieragent}. Codex was used with GPT-5.6 Sol and
GPT-5.6 Terra (max), whereas Claude Code served as the common scaffold for ten
models. Frontier Agent was evaluated with Apodex 1.1 in the Agent Team
configuration.

\paragraph{Models.}
We evaluated twelve frontier models. These comprised GPT-5.6 Sol and GPT-5.6
Terra (max)~\citep{openai2026gpt56}, Grok 4.6~\citep{xai2026grok46}, Kimi
K3~\citep{kimiteam2026k3}, Claude Opus 5~\citep{anthropic2026claudeopus5}, and
Qwen 3.8 Max~\citep{alibaba2026qwen38max}. We also included Qwen3.5-397B-A17B
\citep{qwenteam2026qwen35}, DeepSeek V4 Flash-0731 and DeepSeek V4
Pro-0813~\citep{deepseekai2026deepseekv4}, Apodex
1.1~\citep{apodexteam2026apodex11}, and
GLM-5.2~\citep{zai2026glm52}, together with Gemini 3.7
Flash~\citep{googledeepmind2026gemini37flash} using dynamic thinking. All
models received the same 97 task objectives and task-visible inputs.

\subsection{Task-specific Evaluation and Primary Metrics}\label{sec:grading}

Each task has a task-specific Grader that checks the required files, numerical
results, formats, figures, code execution, and cross-artifact consistency. It
returns a native score $s_{mi}\in[0,100]$ for configuration $m$ on task $i$.
If a rubric-defined semantic criterion is delegated to a Judge, the Judge
remains part of that task's Grader. We use GPT-5.6 Sol as the Judge and run each
Judge-assessed criterion three times.

Let $N=97$ denote the number of evaluated tasks. Because scores from three
Judge passes are averaged, a submission satisfying the complete rubric can be
reported slightly below 100. Following the frozen scoring rule, we therefore
define the full-completion indicator as
$f_{mi}=\mathbb{1}[s_{mi}\geq99.9]$ and report:
\begin{align}
\mathrm{PassRate}_m &= \frac{1}{N}\sum_{i=1}^{N} f_{mi},\\
\mathrm{Avg.\ Score}_m &= \frac{1}{N}\sum_{i=1}^{N} s_{mi}.
\end{align}
Pass Rate is the primary metric and records tasks meeting this strict
full-completion criterion. Avg.\ Score is the mean task score and measures
partial completion. The 99.9 threshold only absorbs minor numerical variation
introduced when the three Judge scores are averaged; it does not relax any
rubric requirement. Because task-specific rubrics are heterogeneous, Avg.\
Score is a descriptive suite-level aggregate rather than a calibrated
cross-task scale.

%% file: sections/experiments.tex


\subsection{Overall performance}

\textbf{Strict completion remained rare across all configurations.} Across
the evaluated systems, Pass Rate ranged from 3.1\% to 20.6\%,
whereas Avg.\ Score ranged from 67.5 to 87.9
(Fig.~\ref{fig:overall-performance} and Table~\ref{tab:overall-results}).
Codex with GPT-5.6 Sol achieved the highest Avg.\ Score of 87.9 and shared the
highest Pass Rate of 20.6\% with Claude Code using Grok 4.6, which obtained an
Avg.\ Score of 86.6. Claude Code with Kimi K3 and Claude Opus 5 each achieved a
Pass Rate of 17.5\%, with Avg.\ Scores of 85.5 and 84.9, respectively.
Codex with GPT-5.6 Terra (max) followed with an Avg.\ Score of 84.7 and a Pass
Rate of 15.5\%.

\begin{center}
  \centering
  \scriptsize
  \captionof{table}{Pass Rate overall and by task difficulty.}
  \label{tab:overall-results}
  \begin{tabular}{@{}llccc@{}}
    \toprule
    \textbf{Model} & \textbf{Agent Scaffold} & \textbf{Overall $\uparrow$} &
    \textbf{Medium $\uparrow$} & \textbf{Hard $\uparrow$} \\
    \midrule
    \modelIcon{openai}\textbf{GPT-5.6 Sol} & \textbf{Codex} & \textbf{20.6\%} & 39.1\% & \textbf{14.9\%} \\
    \modelIcon{xai}\textbf{Grok 4.6} & \textbf{Claude Code} & \textbf{20.6\%} & \textbf{43.5\%} & 13.5\% \\
    \modelIcon{kimi}Kimi K3 & Claude Code & 17.5\% & 39.1\% & 10.8\% \\
    \modelIcon{anthropic}Claude Opus 5 & Claude Code & 17.5\% & 39.1\% & 10.8\% \\
    \modelIcon{openai}GPT-5.6 Terra (max) & Codex & 15.5\% & 39.1\% & 8.1\% \\
    \modelIcon{qwen}Qwen 3.8 Max & Claude Code & 15.5\% & 34.8\% & 9.5\% \\
    \modelIcon{deepseek}DeepSeek V4 Pro-0813 & Claude Code & 13.4\% & 30.4\% & 8.1\% \\
    \modelIcon{deepseek}DeepSeek V4 Flash-0731 & Claude Code & 12.4\% & 34.8\% & 5.4\% \\
    \modelIcon{apodex}Apodex 1.1 & Frontier Agent (Agent Team) & 12.4\% & 30.4\% & 6.8\% \\
    \modelIcon{gemini}Gemini 3.7 Flash & Claude Code & 10.3\% & 21.7\% & 6.8\% \\
    \modelIcon{apodex}Apodex 1.1 & Claude Code & 10.3\% & 30.4\% & 4.1\% \\
    \modelIcon{qwen}Qwen3.5-397B-A17B & Claude Code & 4.1\% & 13.0\% & 1.4\% \\
    \modelIcon{zai}GLM-5.2 & Claude Code & 3.1\% & 4.3\% & 2.7\% \\
    \bottomrule
  \end{tabular}
\end{center}

\begin{center}
  \centering
  \includegraphics[width=\textwidth]{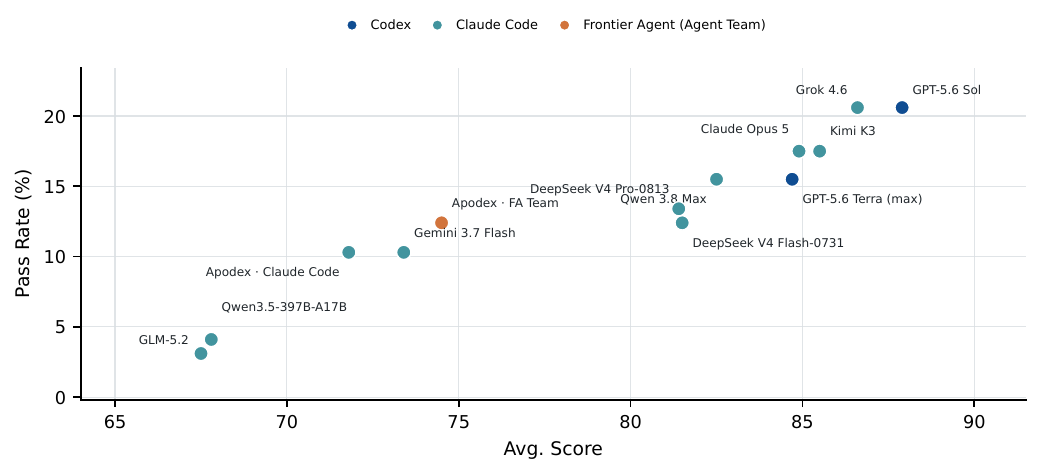}
  \captionof{figure}{Average score versus complete workflow pass rate.}
  \label{fig:overall-performance}
\end{center}

\textbf{High Avg.\ Scores did not imply complete delivery.} Eight configurations
achieved Avg.\ Scores above 80, but none completed more than
20.6\% of tasks under the strict criterion. Thus, high average scores often reflected
artifact bundles that still missed at least one requirement. The two Apodex
1.1 configurations also varied substantially. Frontier Agent (Agent Team)
obtained a higher Avg.\ Score and Pass Rate than Claude Code (74.5 and 12.4\%
versus 71.8 and 10.3\%). This comparison is descriptive and does not isolate a
pure scaffold effect.

\input{sections/heavy_duty_solver}

\subsection{Domain-level performance}

\begin{figure}[H]
  \centering
  \includegraphics[width=\textwidth]{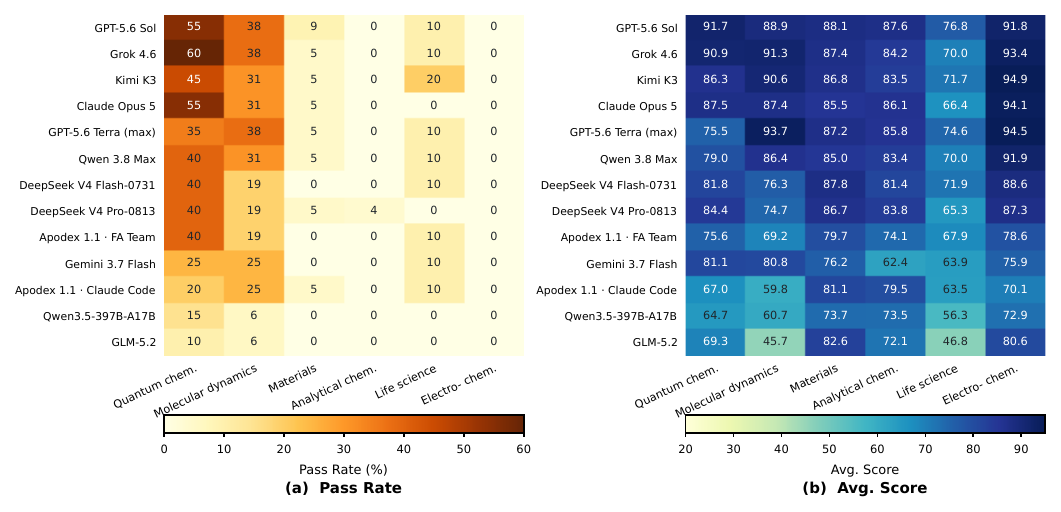}
  \captionsetup{skip=2pt}
  \caption{Domain performance.}
  \label{fig:domain-performance}
\end{figure}

\textbf{Pass Rates were highest in two domains.} Domain-level Avg.\ Score and
Pass Rate were available for every evaluated system
(Fig.~\ref{fig:domain-performance}). In quantum chemistry,
Claude Code with Grok 4.6 achieved the
highest Pass Rate of 60\%, followed by Codex with GPT-5.6 Sol and Claude Code
with Claude Opus 5 at 55\%. In molecular dynamics, GPT-5.6 Sol, GPT-5.6 Terra
(max), and Grok 4.6 each achieved a Pass Rate of 38\%, while Terra obtained the
highest Avg.\ Score of 93.7.

\textbf{High domain-level Avg.\ Scores often coexisted with near-zero Pass Rates.} In
materials characterization, Avg.\ Scores
reached 88.1, but no configuration exceeded a Pass Rate of 9\%. The divergence
was stronger in analytical chemistry and electrochemistry/environment. The
highest
Avg.\ Score in analytical chemistry was 87.6, yet only DeepSeek V4 Pro-0813 completed
any task in that domain, with a Pass Rate of 4\%. Electrochemistry/environment
reached a maximum Avg.\ Score of 94.9, but its
Pass Rate remained 0\% for every configuration. Life-science tasks showed lower
Avg.\ Scores overall; GPT-5.6 Sol achieved the highest Avg.\ Score of 76.8, whereas
Kimi K3 achieved the highest Pass Rate of 20\%.

\textbf{Representative configurations showed distinct domain profiles.}
GPT-5.6 Sol had the broadest leading profile, with the highest Avg.\ Score in
quantum chemistry, materials characterization, analytical chemistry, and life
science. Grok 4.6 achieved the highest quantum-chemistry Pass Rate (60\%) and
shared the highest molecular-dynamics Pass Rate (38\%). The strongest Apodex
1.1 configuration, Frontier Agent (Agent Team), was more competitive in
quantum chemistry (40\% Pass Rate) than in molecular dynamics (19\%) and did
not complete a task in materials characterization, analytical chemistry, or
electrochemistry/environment. Terra's aggregate deficit to Sol
was concentrated in quantum
chemistry: it scored 75.5 versus 91.7 and completed 35\% versus 55\% of tasks,
whereas Terra exceeded Sol in molecular-dynamics Avg.\ Score (93.7 versus
88.9) and electrochemistry/environment Avg.\ Score (94.5 versus
91.8). Gemini 3.7 Flash
also varied across domains: its Avg.\ Score ranged from 62.4 in analytical
chemistry to 81.1 in quantum chemistry, and it completed 25\% of tasks in both
quantum chemistry and molecular dynamics but none in materials
characterization, analytical chemistry, or electrochemistry/environment. These
profiles
show that the aggregate ranking did not capture domain-specific strengths and
weaknesses.
\FloatBarrier

\subsection{Resource use and execution time}

\begin{figure}[H]
  \centering
  \includegraphics[width=\textwidth]{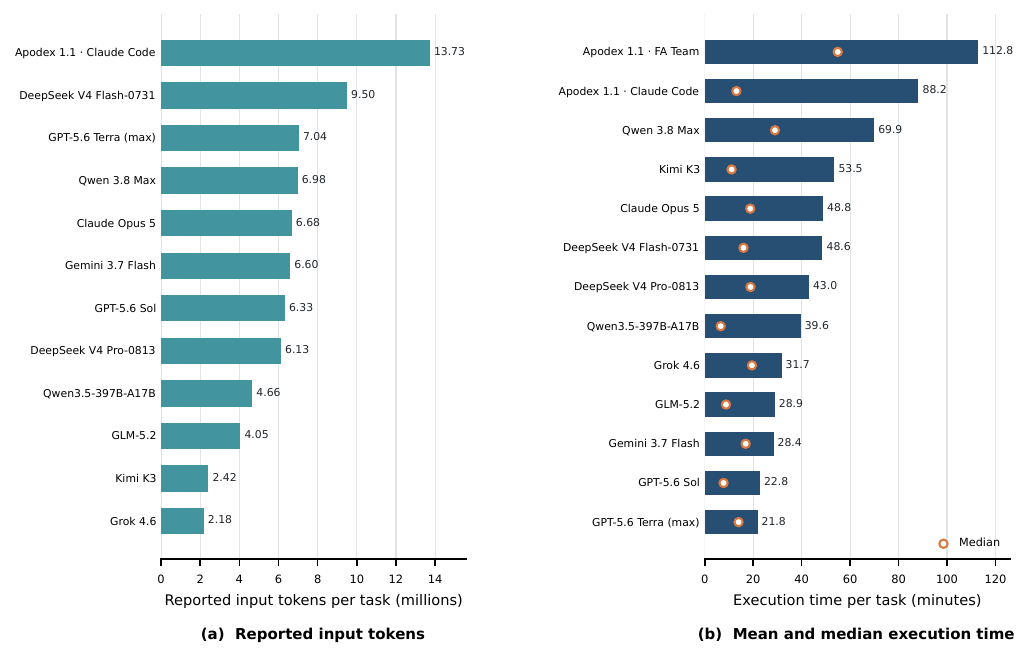}
  \caption{Resource use.}
  \label{fig:resource-performance}
\end{figure}

\textbf{Reported token use varied by more than sixfold.} Among the twelve configurations
with token records, reported input use ranged from 2.183 million tokens per
task for Grok 4.6 with Claude Code to 13.730 million for Apodex 1.1 with Claude
Code. GPT-5.6 Sol used 6.327 million reported input tokens and 23.1 thousand
output tokens per task. GPT-5.6 Terra (max) used 7.039 million input tokens and
38.6 thousand output tokens per task. Their reported cache shares were 98.8\%
and 98.5\%, respectively. Provider-specific tokenizers, caching, and reporting
conventions preclude a hardware-normalized efficiency interpretation. Gemini
3.7 Flash used 6.599 million reported input tokens and 25.8 thousand output
tokens per task, with a 75.3\% cache share (Fig.~\ref{fig:resource-performance}a).

\textbf{Execution times varied substantially and showed long tails.} Across
the evaluated systems, mean execution time ranged from 21.8 to 112.8
minutes per task. Frontier Agent (Agent Team) had the longest mean time at
112.8 minutes, with a median of 54.9 minutes and a 90th percentile of 272.2
minutes. Terra (max) had a higher median than Sol (14.0 versus 7.8
minutes) but a lower mean (21.8 versus 22.8), lower 90th percentile
(35.0 versus 46.1), fewer trajectories over 60 minutes (5 versus 7), and fewer
total machine hours (35.3 versus 36.9), indicating a shorter observed tail.
Gemini 3.7 Flash averaged 28.4 minutes per task, with a median of 16.9 minutes,
a 90th percentile of 71.9 minutes, and 11 trajectories exceeding 60 minutes.
These aggregate summaries describe observed runtimes rather than isolated
model speed.

\FloatBarrier

\subsection{Failure Mode Analysis}

\begin{figure}[t]
  \centering
  \includegraphics[width=\textwidth]{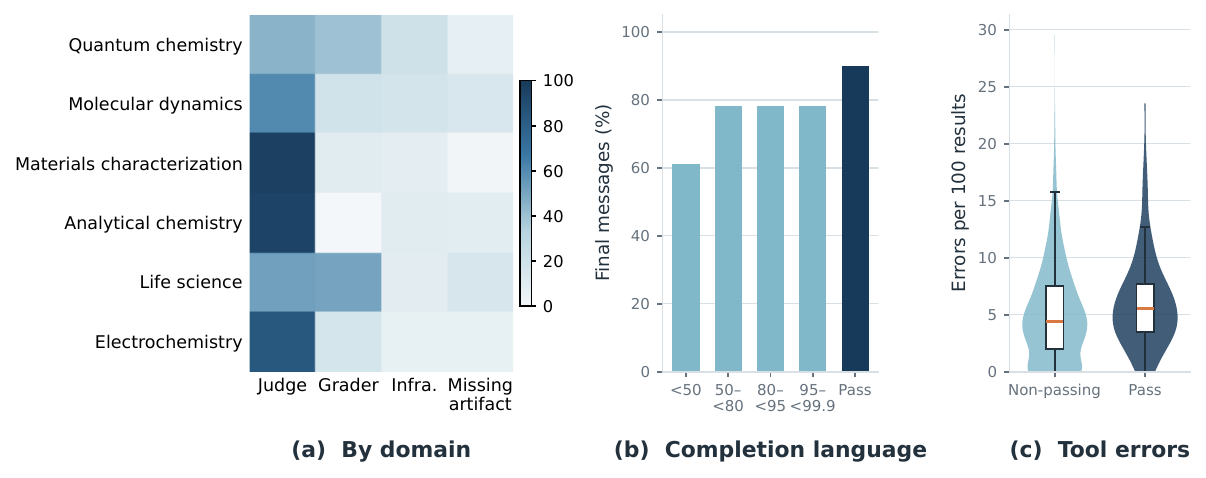}
  \caption{Failure analysis.}
  \label{fig:failure-analysis}
\end{figure}

Figure~\ref{fig:failure-analysis} summarizes incomplete scientific handoffs
from three complementary views: the prevalence and domain distribution of
machine-observable contract-breach signatures, completion language in final
messages, and raw tool errors. The signatures can overlap, and the trajectory
comparisons are restricted to Claude Code, for which a common event schema was
available across models. These analyses are descriptive and do not assign a
unique cause to any run.

\textbf{Failure signatures varied substantially by domain.} Judge-assessed
artifact shortfalls appeared
in 97\% of non-passing materials-characterization submissions, 95\% of
analytical-chemistry submissions, and 85\% of electrochemistry/environment
submissions,
compared with 43\% in quantum chemistry. Deterministic Grader diagnostics were
more common in life science (50\%) and quantum chemistry (40\%) than in the
other domains (Fig.~\ref{fig:failure-analysis}a). These contrasts partly reflect differences in task contracts
and evaluator composition and therefore do not establish intrinsic
domain-specific causes.

\textbf{Completion language did not reliably indicate successful delivery.}
Among the 970 Claude Code trajectories, 641 of the 849 non-passing runs
(75.5\%) had a final
message containing lexical completion language; only 13 (1.5\%)
explicitly indicated that work was still running, waiting, or in progress.
Completion language was
less frequent among runs scoring below 50 (61.2\%) but remained above 78\% in
each non-passing band above 50; it appeared in 90.1\% of passing runs
(Fig.~\ref{fig:failure-analysis}b). Final self-reporting is therefore not a
reliable substitute for evaluating the submitted artifact bundle.

\textbf{Raw tool errors did not predict success.}
At least one tool error occurred in 80.7\% of non-passing and 94.2\% of passing
Claude Code trajectories. Median error rates were 4.4 and 5.6 per 100 tool
results, respectively (Fig.~\ref{fig:failure-analysis}c). Successful
trajectories can encounter and recover from tool errors; the presence of an
error alone does not identify the eventual contract breach.

\FloatBarrier

%% file: sections/heavy_duty_solver.tex
\subsection{Heavy Duty Solver Capability Assessment}\label{sec:hds6}

\paragraph{Protocol and coverage.}
Following the HDS6 evaluation framework introduced in \emph{Apodex
Discovery}~\citep{wang2026apodexdiscovery}, we assess six Heavy Duty Solver
capabilities independently of final-task success. We apply this framework
through the archived FrontierChallenge HDS6 campaign from September 7, 2026. This is a retrospective process assessment of frozen trajectories,
complementing the artifact-based completion metrics in
Section~\ref{sec:grading}. It covers 12 model--scaffold configurations on all
97 tasks (1,164 scorecards). Apodex 1.1 is included only with Claude Code:
the Frontier Agent runs lack the structured ATIF trajectories required by
this assessment. The partially covered GLM-5.3 configuration is also excluded.
These results therefore do not constitute an HDS6 assessment of every scaffold
in the main benchmark.

The six dimensions are \emph{Coherence} (long-horizon state and cross-artifact
consistency), \emph{Evidence} (fidelity to executed observations),
\emph{Alternatives} (hypothesis and method management), \emph{Scope}
(boundary and failure reasoning), \emph{Tools} (tool use and execution-state
management), and \emph{Repair} (self-correction under verification).
Each task uses 13 shared rubric items plus 6--10 task-specific items with
0/1/2 anchors; the archived scorer reports overall and capability scores on
a 1--4 base scale. A reviewed opportunity matrix determines which
capabilities are applicable. Missing opportunities are excluded rather than
assigned zero: Alternatives and Repair are marked not applicable on 47 tasks
each. The judge consumes converted trajectories and task packs, without direct
access to the verifier's reference outputs. Private reasoning channels are
stripped for all configurations. The archived TERRA panel uses GPT-5.6 Terra
as judge and reviewer, GPT-5.6 Luna as sweeper, and Claude Opus 5 as arbiter.

\paragraph{Aggregation.}
Let $A_m$ be the available runs for configuration $m$, and $V_m\subseteq A_m$
its graded runs passing the HDS6 integrity gate. This gate is distinct from
the benchmark's task-completion threshold. For archived overall score
$h_{mi}$, we reproduce the campaign's coverage-adjusted score:
\begin{equation}
H_m = \left(\frac{1}{|V_m|}\sum_{i\in V_m}h_{mi}\right)
      \sqrt{\frac{|V_m|}{|A_m|}}.
\end{equation}
Tasks receive equal weight. For each capability, we average its available
scores within $V_m$ and apply the same configuration-level multiplier.
Gate failures remain in $A_m$ but do not contribute to the score mean.
Five Apodex 1.1 runs containing only an API authentication failure have no
process grade and are removed from both sets, giving $|A_m|=92$ for that
configuration and 97 for every other configuration. The original outcome
metrics retain their original 97-task denominator. Coverage adjustment can
reduce a score below the base scale's lower bound; it is not a pass rate.

\input{tables/hds6_results}

\paragraph{Results and interpretation.}
The HDS6 ranking broadly agrees with the outcome-based ranking reported
above. Across the 12 configurations, HDS6 correlates strongly with Avg.\ Score
(Pearson $r=0.874$, Spearman $\rho=0.762$) and Pass Rate ($r=0.828$,
$\rho=0.799$). The association also holds within individual tasks: the mean
Pearson correlation between process score and task score is 0.475 across
94 eligible tasks, with positive correlations on 85\% of them. Because HDS6
assesses agents' actions and their observed results independently of the
artifact grader, this agreement provides evidence that success on
\benchmark{} reflects Heavy Duty Solver capabilities. The benchmark's
scientific workflows require agents to sustain coherent execution, ground
outputs in evidence, manage tools, and verify their work, making
\benchmark{} a substantive test of these capabilities.

Table~\ref{tab:hds6-results} reveals both overall performance differences
and specific capability profiles. Claude Opus 5 achieves the highest HDS6
score (3.768), followed by GPT-5.6 Sol (3.670), Qwen 3.8 Max (3.624), and
GPT-5.6 Terra (3.612). Evidence is the highest or nearly highest dimension
for every configuration, whereas Scope is consistently the lowest
(2.185--3.508), identifying boundary checks and failure reasoning as a common
weakness. Apodex 1.1 with Claude Code scores 3.293, above Qwen3.5-397B-A17B
(2.643), GLM-5.2 (3.152), and Gemini 3.7 Flash (3.016), but below Kimi K3
(3.496) and DeepSeek V4 Flash-0731 (3.506). Its largest gaps relative to
Claude Opus 5 are in Scope (2.853 versus 3.508) and Alternatives
(3.262 versus 3.823), highlighting boundary/failure reasoning and hypothesis
management as concrete targets for improvement.

%% file: tables/hds6_results.tex
\begin{table}[t]
\centering
\scriptsize
\caption{Exploratory Heavy Duty Solver (HDS6) process assessment. Scores are coverage-adjusted; higher is better. C: Coherence; E: Evidence; A: Alternatives; S: Scope; T: Tools; R: Repair. V/N denotes gate-passing graded runs / available runs.}
\label{tab:hds6-results}
\setlength{\tabcolsep}{3pt}
\resizebox{\textwidth}{!}{%
\begin{tabular}{@{}llrrrrrrrr@{}}
\toprule
Model & Scaffold & HDS6 & C & E & A & S & T & R & V/N \\
\midrule
Claude Opus 5 & Claude Code & 3.768 & 3.890 & 3.951 & 3.823 & 3.508 & 3.704 & 3.653 & 97/97 \\
GPT-5.6 Sol & Codex & 3.670 & 3.874 & 3.919 & 3.431 & 3.341 & 3.641 & 3.533 & 96/97 \\
Qwen 3.8 Max & Claude Code & 3.624 & 3.845 & 3.842 & 3.472 & 3.186 & 3.601 & 3.607 & 96/97 \\
GPT-5.6 Terra (max) & Codex & 3.612 & 3.818 & 3.876 & 3.413 & 3.184 & 3.582 & 3.560 & 95/97 \\
Grok 4.6 & Claude Code & 3.562 & 3.777 & 3.853 & 3.436 & 3.196 & 3.523 & 3.338 & 92/97 \\
DeepSeek V4 Pro-0813 & Claude Code & 3.536 & 3.787 & 3.857 & 3.367 & 3.003 & 3.536 & 3.427 & 97/97 \\
DeepSeek V4 Flash-0731 & Claude Code & 3.506 & 3.774 & 3.767 & 3.226 & 3.056 & 3.488 & 3.434 & 92/97 \\
Kimi K3 & Claude Code & 3.496 & 3.776 & 3.815 & 3.316 & 2.996 & 3.523 & 3.252 & 95/97 \\
Apodex 1.1 & Claude Code & 3.293 & 3.460 & 3.544 & 3.262 & 2.853 & 3.304 & 3.300 & 82/92 \\
GLM-5.2 & Claude Code & 3.152 & 3.302 & 3.532 & 3.176 & 2.652 & 3.232 & 2.950 & 89/97 \\
Gemini 3.7 Flash & Claude Code & 3.016 & 3.258 & 3.451 & 2.854 & 2.438 & 3.098 & 2.499 & 90/97 \\
Qwen3.5-397B-A17B & Claude Code & 2.643 & 2.822 & 2.998 & 2.341 & 2.185 & 2.683 & 2.410 & 69/97 \\
\bottomrule
\end{tabular}}
\end{table}

%% file: sections/conclusion.tex
\section{Conclusion}\label{sec:conclusion}

\benchmark{} evaluates whether scientific agents can complete specified,
multi-stage workflows and deliver mutually consistent artifacts rather than
merely produce plausible answers. The benchmark comprises 300 workflows; this
study releases and evaluates 97 across six scientific domains using
task-specific executable Graders. Across the evaluated models and agent
scaffolds, Pass Rate ranged from 3.1\% to 20.6\%, despite Avg.\ Scores of
67.5 to 87.9. GPT-5.6 Sol with Codex achieved the highest Avg.\ Score and shared
the highest Pass Rate with Grok 4.6 using Claude Code. Partial progress
translated especially poorly into complete delivery in analytical chemistry
and electrochemistry/environment, and the distinct domain
profiles show that aggregate rankings do not capture every scientific setting.
Failure analysis further
showed that 75.5\% of non-passing Claude Code trajectories ended with
completion language, while tool errors were frequent in both passing and
non-passing runs. Final self-reports and the mere presence of errors are
therefore weak indicators of successful delivery. The findings are limited to
the released task set, evaluated configurations, Claude Code trajectory scope,
provider-specific resource accounting, and single runs. Within these bounds,
the results establish complete, contract-level delivery as a distinct and
unresolved capability. More reliable scientific agents will require explicit
contract tracking, cross-artifact validation, and evidence-based completion
checks.

%% file: sections/appendix.tex
\section{Contributors}\label{app:contributors}

Liangcai Su\textsuperscript{*}, Zhaopeng Feng\textsuperscript{*}, Zhuo
Chen\textsuperscript{*}, Zhen Zhang\textsuperscript{*}, Xiang Lin, Ruilin Li, Handuo Zhang, Ning Wang,
Kailong Wen, Yueqi Guo, Feng Xing, Yiling Guo, Brian Wang, Chenxiong Qian, Simon Shaolei Du,
Lidong Bing, and Xinyu
Wang\textsuperscript{\textdagger}.

{\footnotesize\textsuperscript{*}Contributed equally;
\textsuperscript{\textdagger}Project lead.}

\input{sections/illustrative_task_cases}

\section{Acknowledgements}\label{app:acknowledgements}

We thank all experts who contributed to task authoring and proofreading. We
are especially grateful for their careful reviews and detailed corrections,
which improved the clarity, scientific accuracy, and evaluability of the
tasks.

%% file: sections/illustrative_task_cases.tex
\section{Illustrative Task Case}\label{app:case-studies}

The following three cases illustrate the benchmark unit at the level visible to
an evaluated system: a scientific objective, a frozen input bundle, a required
workflow, and an artifact contract.  They intentionally disclose neither model
submissions nor scores, evaluator code, hidden checks, expected outputs, or
reference artifacts.  Accordingly, the examples characterize what must be
done and handed off, not the answer to any task.

\subsection{Cell-migration wound-healing assay}

\paragraph{Task and visible evidence.}
This case is an image-analysis and statistical workflow built from 18 supplied
bright-field microscopy images.  The frozen input crosses two groups (control
and experimental), three time points (0, 12, and 24 hours), and three biological
replicates.  In every image, the experimenter has already marked the closed
boundary of the unmigrated region in white.  Figure~\ref{fig:case-migration}
shows one complete time course from each group exactly as provided to the
system; the outlines are part of the source images, not generated results.

\begin{center}
  \begin{minipage}{0.96\textwidth}
    \centering
    \raggedright\footnotesize\textbf{Control, replicate 1}\par\vspace{2pt}
    \centering
    \begin{minipage}[t]{0.318\linewidth}\centering
      \includegraphics[width=\linewidth]{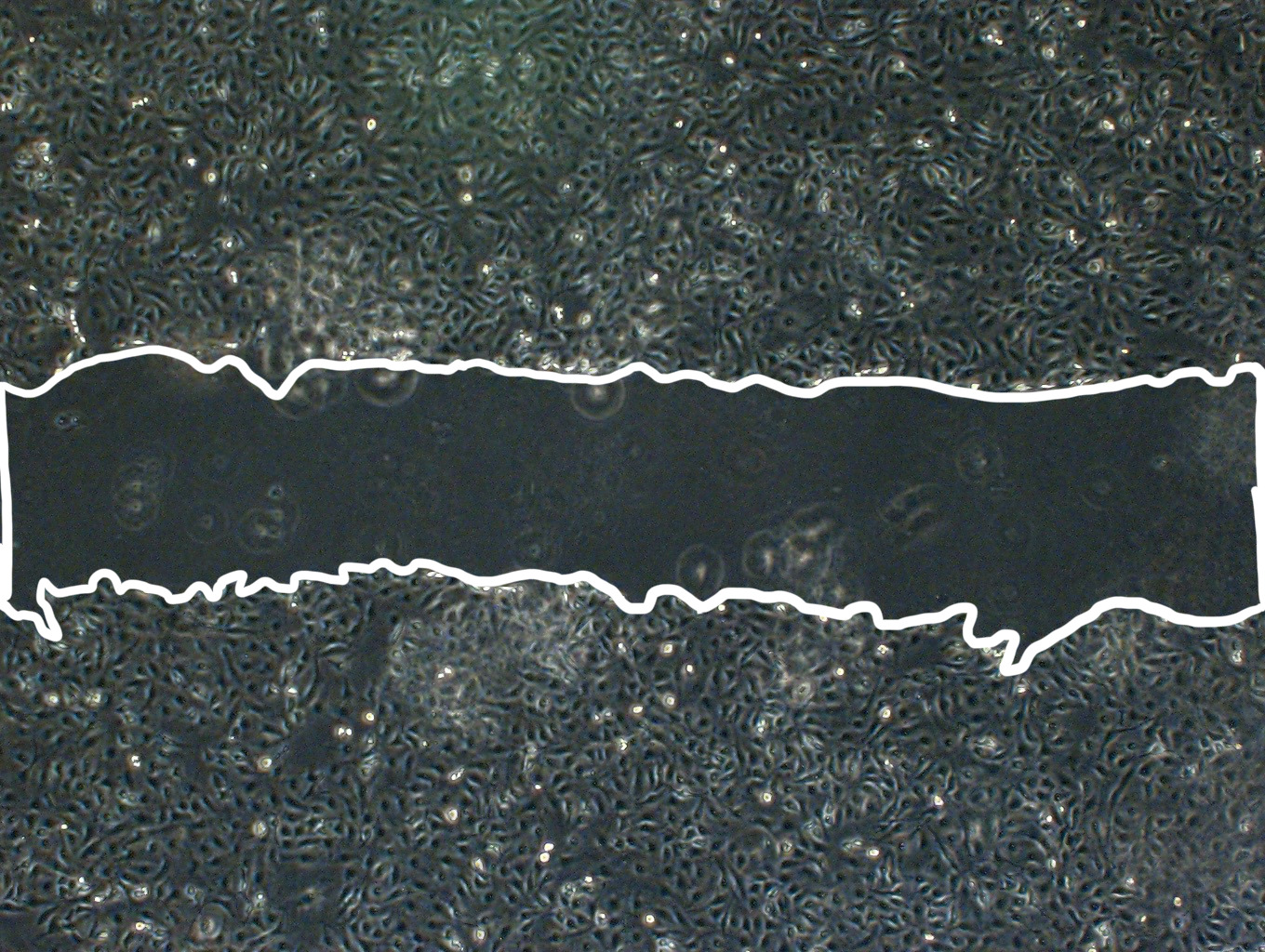}\\[-0.2ex]
      \footnotesize\textbf{0 h}
    \end{minipage}\hfill
    \begin{minipage}[t]{0.318\linewidth}\centering
      \includegraphics[width=\linewidth]{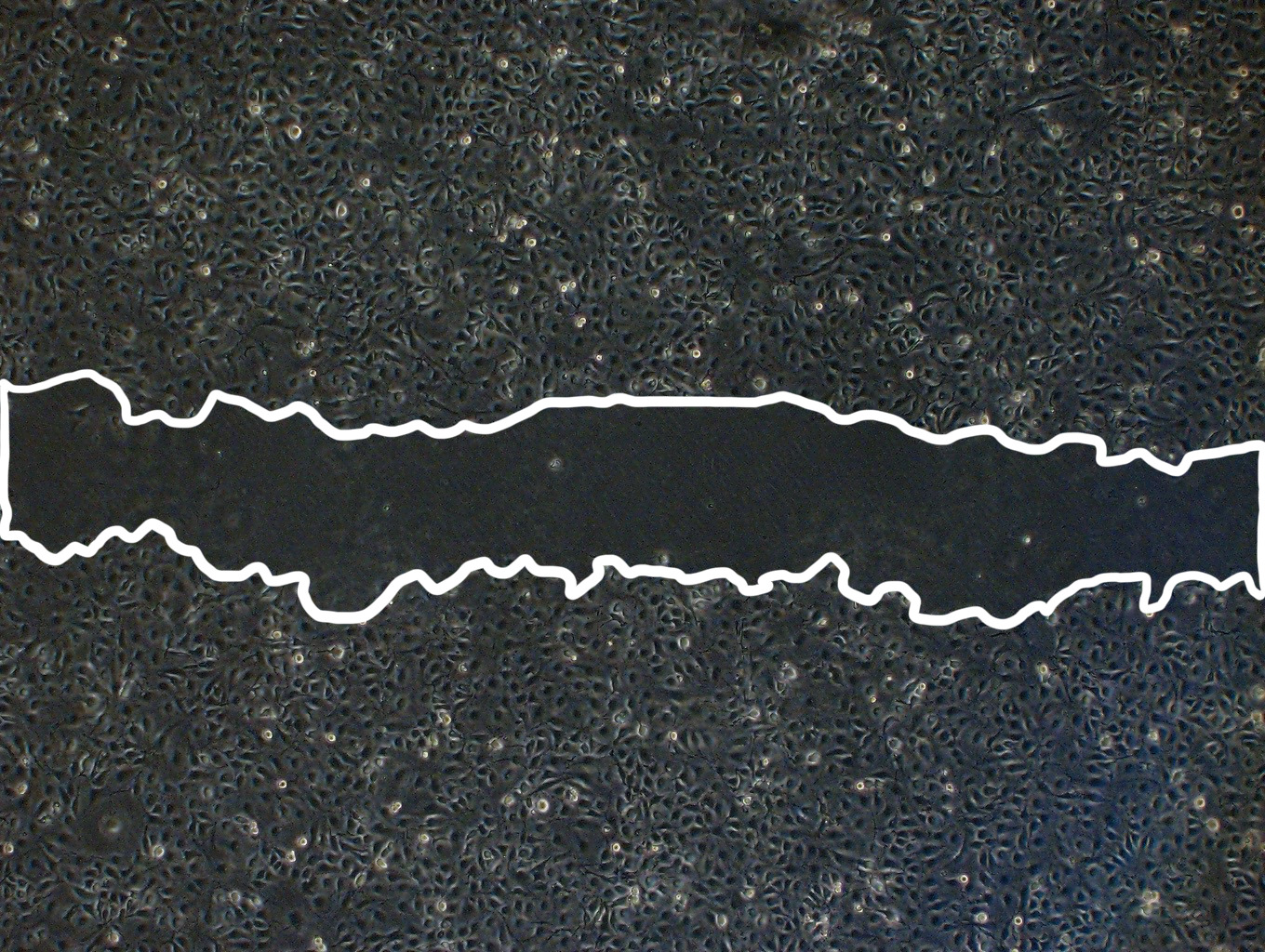}\\[-0.2ex]
      \footnotesize\textbf{12 h}
    \end{minipage}\hfill
    \begin{minipage}[t]{0.318\linewidth}\centering
      \includegraphics[width=\linewidth]{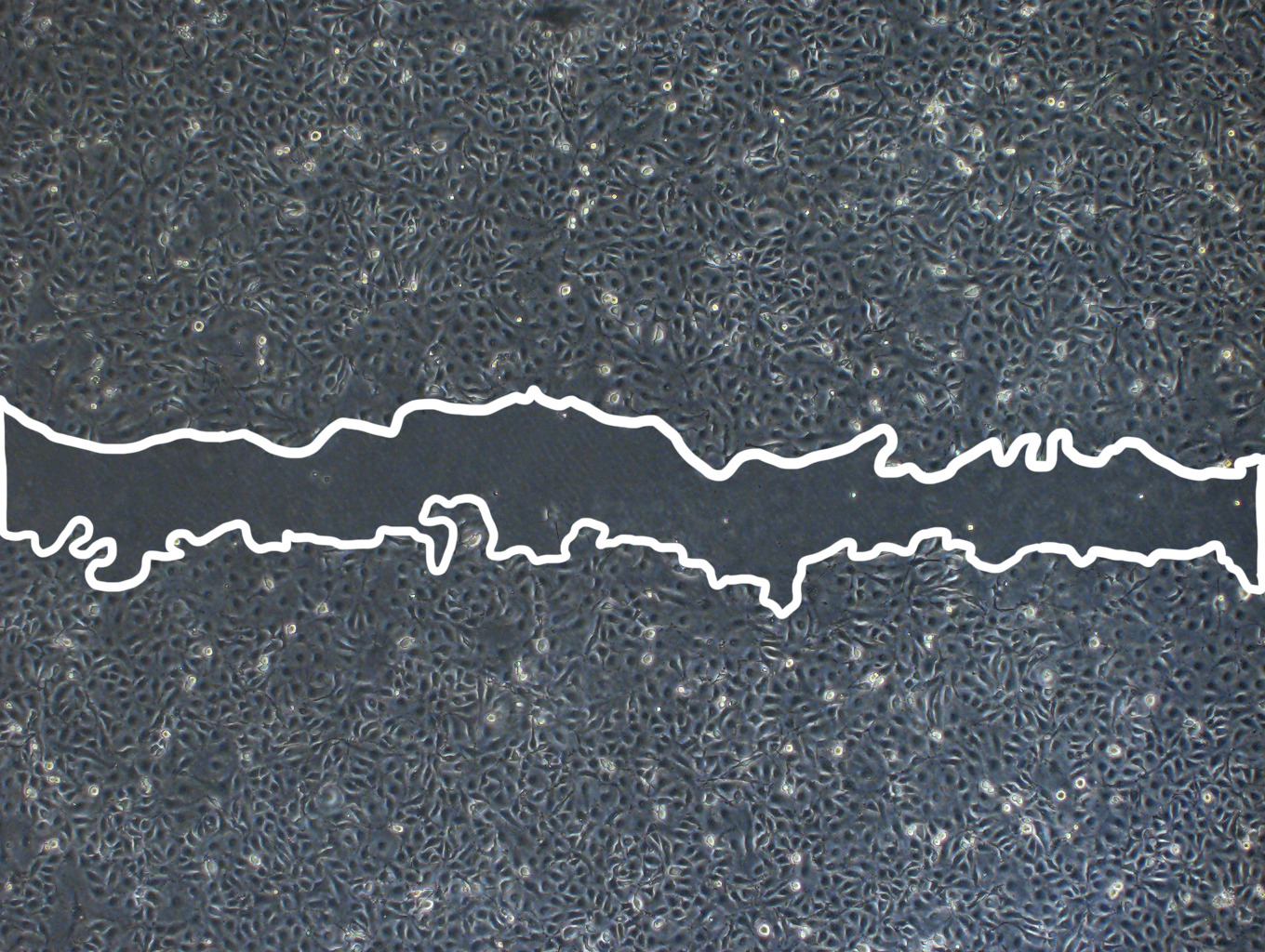}\\[-0.2ex]
      \footnotesize\textbf{24 h}
    \end{minipage}

    \vspace{5pt}
    \raggedright\footnotesize\textbf{Experimental, replicate 1}\par\vspace{2pt}
    \centering
    \begin{minipage}[t]{0.318\linewidth}\centering
      \includegraphics[width=\linewidth]{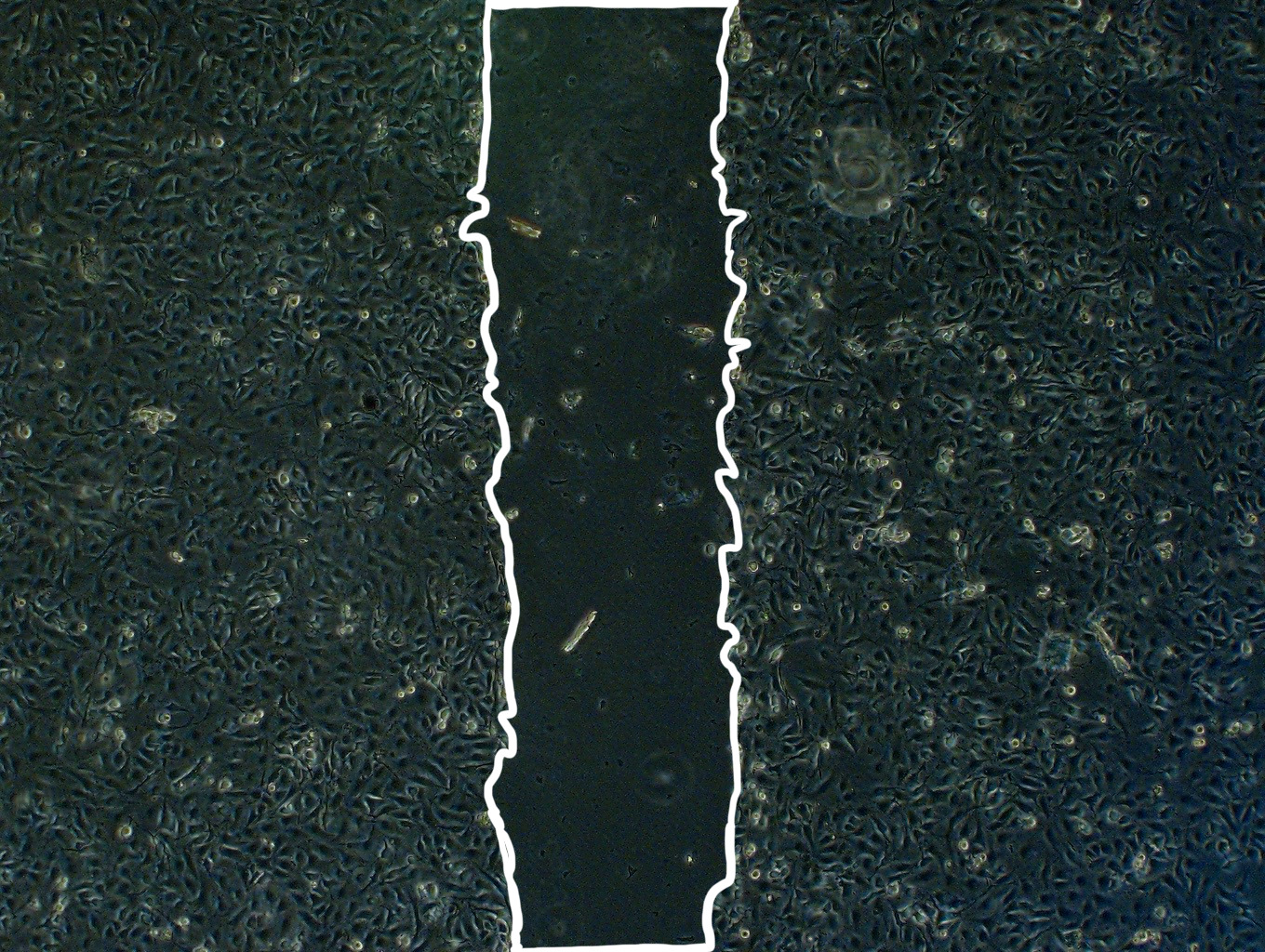}\\[-0.2ex]
      \footnotesize\textbf{0 h}
    \end{minipage}\hfill
    \begin{minipage}[t]{0.318\linewidth}\centering
      \includegraphics[width=\linewidth]{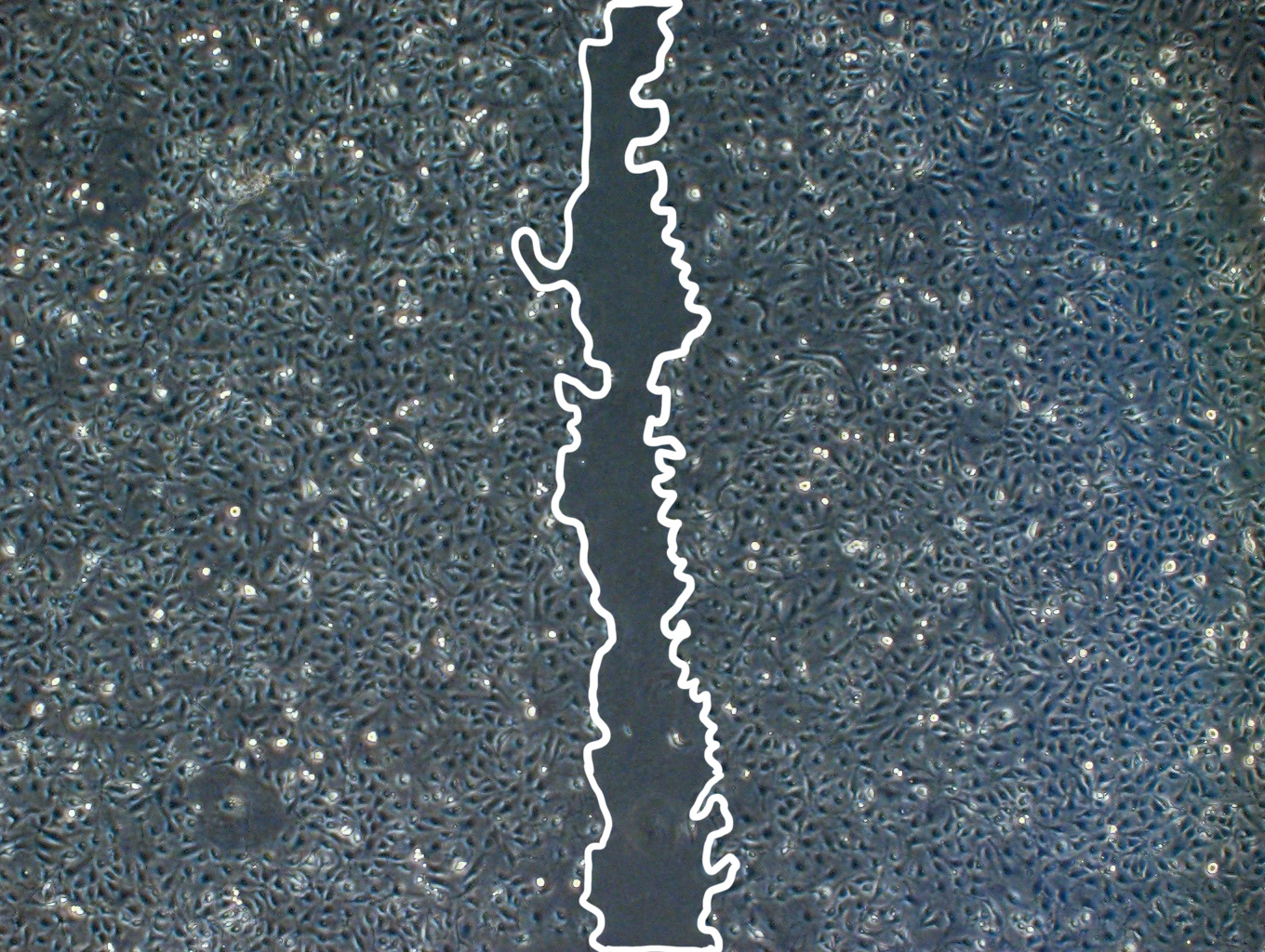}\\[-0.2ex]
      \footnotesize\textbf{12 h}
    \end{minipage}\hfill
    \begin{minipage}[t]{0.318\linewidth}\centering
      \includegraphics[width=\linewidth]{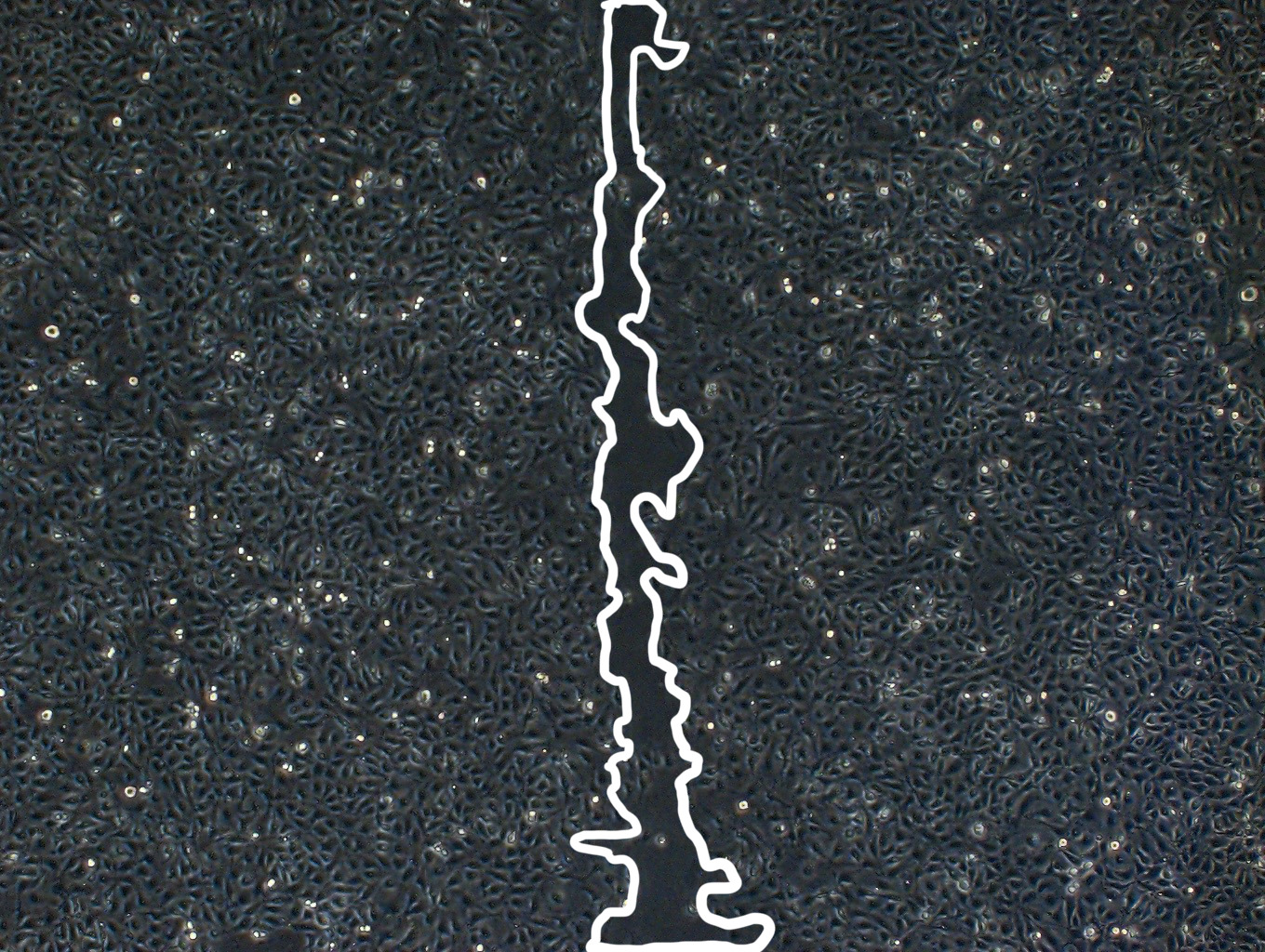}\\[-0.2ex]
      \footnotesize\textbf{24 h}
    \end{minipage}
    \captionof{figure}{Agent-visible input from the wound-healing case.  Columns
    are the three supplied time points; rows show one biological replicate from
    each group.  White contours were drawn by the experimenter in the original
    images to delineate the unmigrated region.  The remaining twelve images
    supply two additional replicates for every group--time combination.}
    \label{fig:case-migration}
  \end{minipage}
\end{center}

\paragraph{Expected handoff.}
The system must segment the marked unmigrated region in all 18 images, inspect
segmentation quality, pair time points by group and replicate, and compute the
12-hour and 24-hour migration rates relative to the matching 0-hour image.  It
then performs separate between-group tests at both follow-up times.  The
checkable handoff consists of a reproducible analysis script, an image-level
table, a group-statistics table, two labeled summary plots, and a methodological
report.  No measured area, migration rate, test statistic, or conclusion is
shown here.

\subsection{TLC monitoring of a Suzuki coupling}

\paragraph{Task and visible evidence.}
This case is an analytical-chemistry image workflow centered on a supplied
UV254 thin-layer chromatography plate.  Its eight lanes comprise starting-
material and product standards, a co-spot, and reaction samples collected at
0, 15, 30, 60, and 120 minutes.  The frozen input also supplies the plate map,
compound-reference Rf windows and response factors, sampling metadata, image
geometry, integration settings, and the endpoint rule.  The spotting line and
solvent front are intentionally tilted, so Rf must be computed from the local
geometry at each lane.

\paragraph{Expected handoff.}
The system must detect and integrate spots after local background correction,
assign compounds using the standards and co-spot, calculate lane-specific Rf
values, apply the declared response correction, reconstruct reaction progress,
and recommend an endpoint under the supplied rule.  Its handoff includes a
reproducible script; plate-QC, spot-level, lane-composition, reaction-progress,
and endpoint tables; an annotated plate; lane profiles; and a report.  Figure~
\ref{fig:case-tlc} is the unmodified agent-visible plate image and contains no
derived annotations, integrated intensities, conversion values, or endpoint.

\begin{center}
  \begin{minipage}{0.96\textwidth}
    \centering
    \begin{minipage}[c]{0.42\linewidth}
      \centering
      \includegraphics[width=\linewidth]{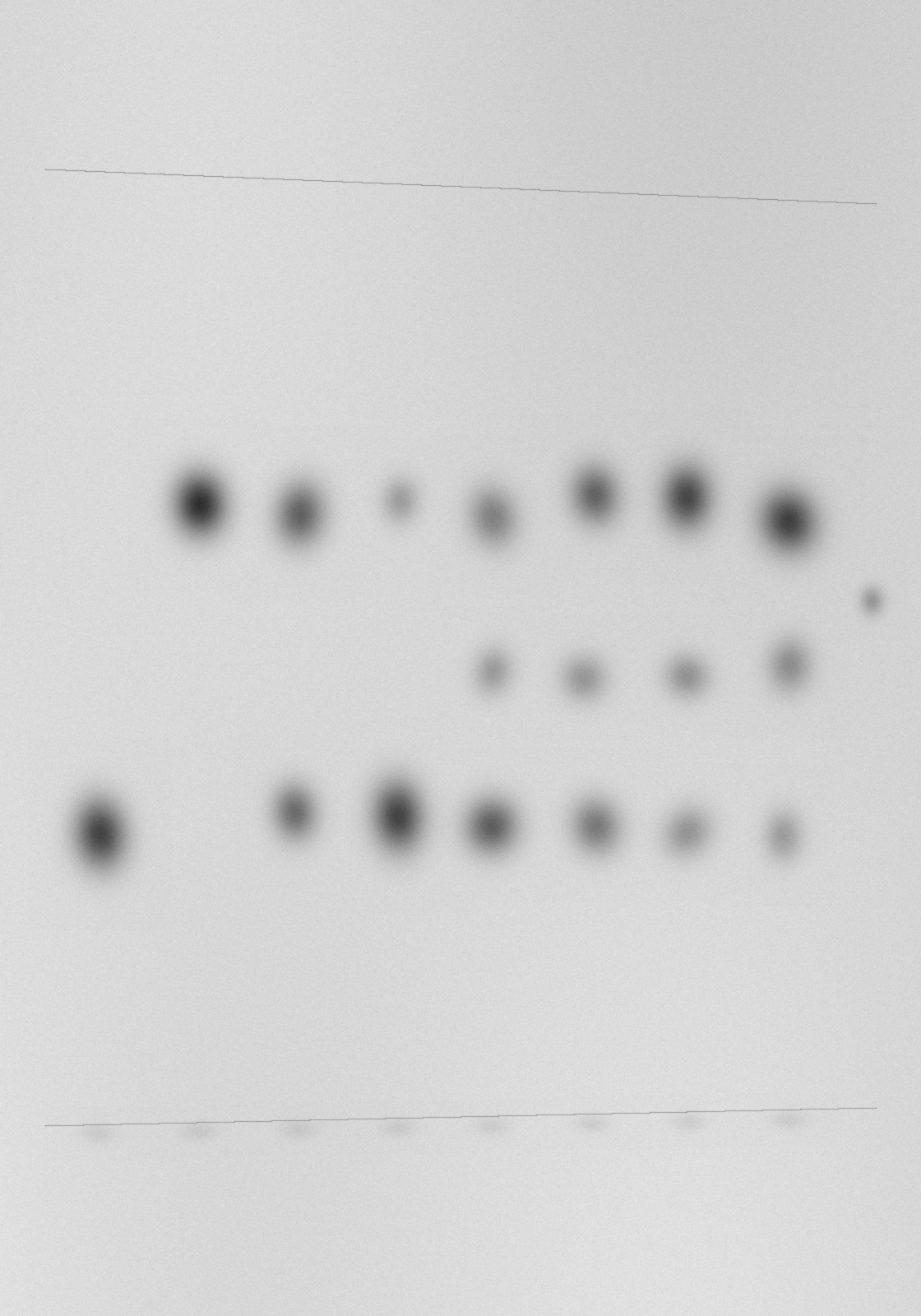}
    \end{minipage}\hfill
    \begin{minipage}[c]{0.51\linewidth}
      \small
      \textbf{Agent-visible plate map}\par\vspace{4pt}
      \setlength{\tabcolsep}{4pt}
      \renewcommand{\arraystretch}{1.12}
      \begin{tabular}{@{}lll@{}}
        \toprule
        \textbf{Lane} & \textbf{Sample} & \textbf{Time} \\
        \midrule
        L01 & Starting-material standard & -- \\
        L02 & Product standard & -- \\
        L03 & Co-spot & -- \\
        L04 & Reaction sample & 0 min \\
        L05 & Reaction sample & 15 min \\
        L06 & Reaction sample & 30 min \\
        L07 & Reaction sample & 60 min \\
        L08 & Reaction sample & 120 min \\
        \bottomrule
      \end{tabular}

      \vspace{9pt}
      \footnotesize
      Dark UV-quenching spots are the measured image signal.  Lane centers,
      approximate line endpoints, compound search windows, response factors,
      and the endpoint criterion are supplied separately as structured input.
    \end{minipage}
    \captionof{figure}{Unmodified agent-visible TLC input for the Suzuki-
    coupling case, shown with the supplied lane map.  The plate contains
    standards, a co-spot, and a five-time-point reaction series.  No detected
    spot boundary, Rf assignment, corrected intensity, composition, or endpoint
    decision is displayed.}
    \label{fig:case-tlc}
  \end{minipage}
\end{center}

\subsection{Reaction-calorimetry safety assessment}

\paragraph{Task and visible evidence.}
This case is a hard process-chemistry and thermal-safety workflow.  Its frozen
input contains electrical calibration pulses, blank-dosing experiments,
reaction runs, run and protocol metadata, and a configuration containing the
screening rules.  The left side of Figure~\ref{fig:case-calorimetry} plots three
representative raw input traces exactly as supplied; no baseline correction,
integration, derived metric, or protocol decision is displayed.

\paragraph{Expected handoff.}
The analysis must calibrate and correct heat flow, derive run-level metrics,
summarize replicates by protocol, and apply the declared screening criteria.
A reproducible script must produce mutually traceable calibration, correction,
time-series, run-metric, protocol-summary, and decision tables, together with
diagnostic figures and a report.  The right side of Figure~\ref{fig:case-calorimetry}
shows this audit chain rather than its outcome.

\begin{center}
  \begin{minipage}{0.96\textwidth}
    \centering
    \includegraphics[width=\linewidth]{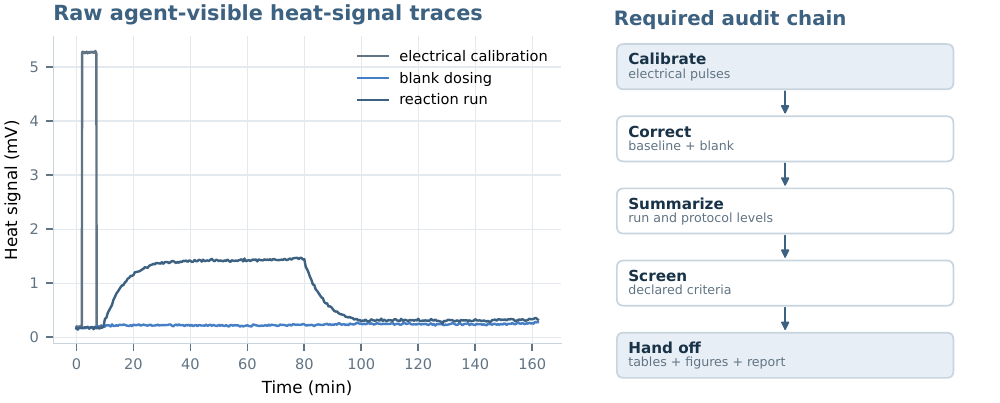}
    \captionof{figure}{Answer-free view of the reaction-calorimetry case.  Raw
    agent-visible signal traces provide visual context, while the schematic
    records the required path from calibration evidence to a complete handoff.}
    \label{fig:case-calorimetry}
  \end{minipage}
\end{center}